\documentclass[twoside,leqno,twocolumn]{article}

\usepackage[letterpaper]{geometry}
\usepackage{float}
\usepackage{ltexpprt}
\usepackage{hyperref}
\usepackage{cite}
\usepackage{amsmath,amssymb,amsfonts}
\usepackage{algorithmic}
\usepackage{graphicx}
\usepackage{textcomp}
\usepackage{xcolor}
\usepackage{caption}
\usepackage{booktabs}
\usepackage{subcaption}
\usepackage{algorithm}
\usepackage{float}
\usepackage{array}
\newcommand{\hide}[1]{}

\providecommand{\keywords}[1]
{
  \small	
  \textbf{\textit{Keywords---}} #1
}
\newcommand{\yq}{\mathcal{Y}_{\tau}^q}
\newcommand{\ys}{\mathcal{Y}_{\tau}^s}
\newcommand{\xq}{\mathcal{X}_{\tau}^q}
\newcommand{\xs}{\mathcal{X}_{\tau}^s}
\newcommand{\cq}{\mathcal{C}_{\tau}^q}
\newcommand{\cs}{\mathcal{C}_{\tau}^s}
\newcommand{\zq}{\mathcal{Z}_{\tau}^q}
\newcommand{\zs}{\mathcal{Z}_{\tau}^s}
\newcommand{\pt}{\theta_{\tau}}
\begin{document}

\newcommand\relatedversion{}

\title{\Large Concept Discovery for Fast Adaptation\relatedversion}
\author{Shengyu Feng\thanks{Language Technology Institute, Carnegie Mellon University. Email: shengyuf@andrew.cmu.edu.}
\and Hanghang Tong\thanks{Department of Computer Science, University of Illinois at Urbana-Champaign. Email: htong@illinois.edu.}}

\date{}

\maketitle


\fancyfoot[R]{\scriptsize{Copyright \textcopyright\ 2023 by SIAM\\
Unauthorized reproduction of this article is prohibited}}





\begin{abstract} 
The advances in deep learning have enabled machine learning methods to outperform human beings in various areas, but it remains a great challenge for a well-trained model to quickly adapt to a new task. One promising solution to realize this goal is through meta-learning, also known as learning to learn, which has achieved promising results in few-shot learning. However, current approaches are still enormously different from human beings' learning process, especially in the ability to extract structural and transferable knowledge. This drawback makes current meta-learning frameworks non-interpretable and hard to extend to more complex tasks. We tackle this problem by introducing concept discovery to the few-shot learning problem, where we achieve more effective adaptation by meta-learning the structure among the data features, leading to a composite representation of the data. Our proposed method \textit{Concept-Based Model-Agnostic Meta-Learning} (COMAML) has been shown to achieve consistent improvements in the structured data for both synthesized datasets and real-world datasets. 
\end{abstract}

\section{Introduction}
Deep neural networks are known for their strong expressivity  in modeling high-dimensional data and have achieved great success in various domains. But deep neural networks are notoriously known to be data-hungry, usually requiring a bunch of labeled data for training. By contrast, human beings are good at learning new tasks through only a few trials, which reveals the lack of this human intelligence in the current deep learning methods.

In machine learning, few-shot learning \cite{855856, 1597116, Koch2015SiameseNN} is the task to tackle this weakness where the model needs to generalize to a new task through training on limited samples.  The fundamental challenge of few-shot learning lies in the risk of overfitting on the training samples, and meta-learning deals with it by explicitly learning a model that can generalize well on the new task with minimum adaptation.  The recent advances in meta-learning have achieved great breakthroughs in various fields such as image classification \cite{finn2017modelagnostic, lee2019metalearning, triantafillou2020metadataset}, robot manipulation \cite{yu2018oneshot, gupta2018metareinforcement, yu2021metaworld}, natural language processing \cite{gu2018metalearning, schick-schutze-2021-exploiting} and graph learning \cite{zugner2018adversarial, franceschi2020learning, yao2020graph}.

\begin{figure}[h]
\centering
\begin{subfigure}[h]{\linewidth}
\centering
\includegraphics[width=.6\linewidth]{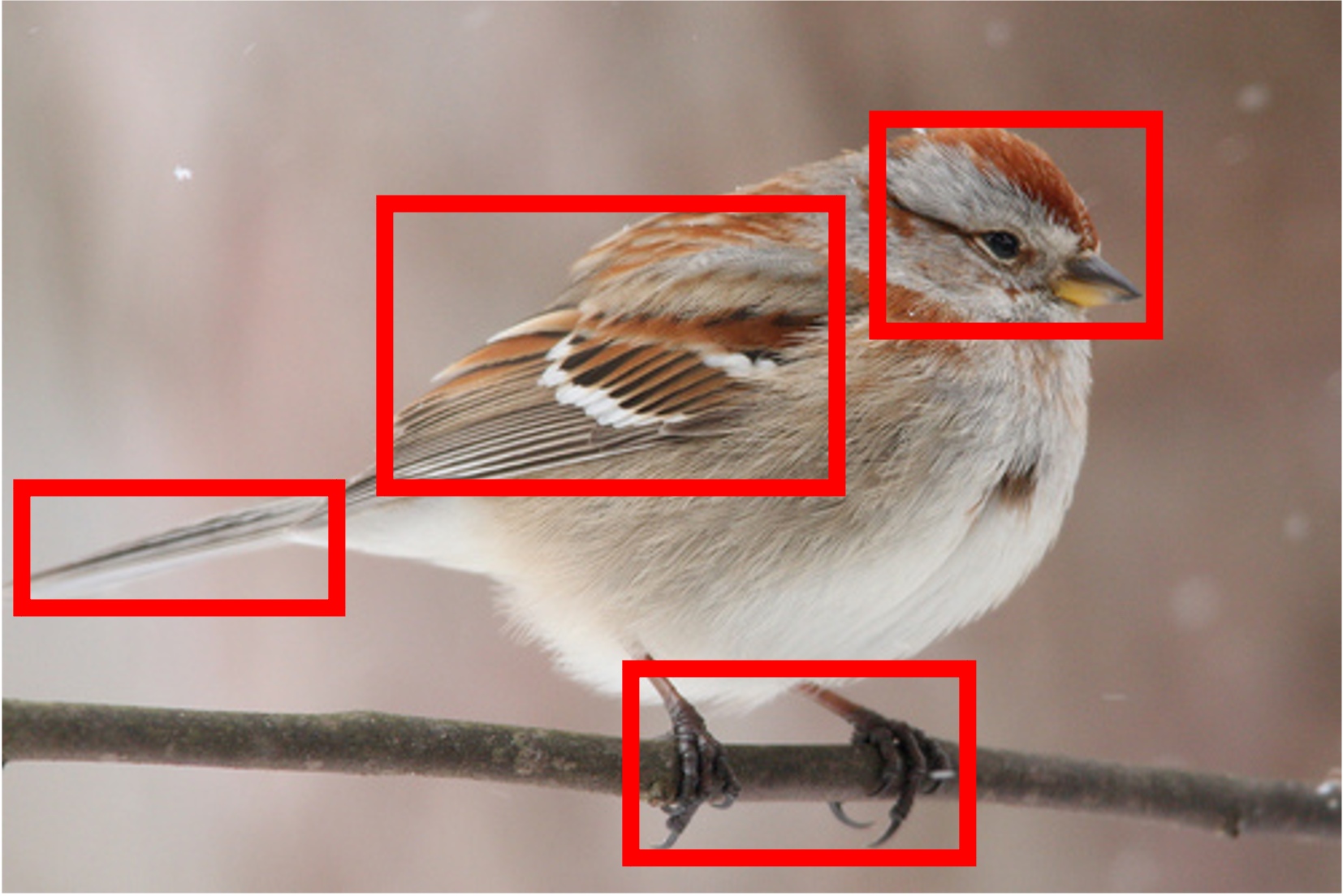}
\caption{Observed data with concepts annotated}
\label{fig:bird}
\end{subfigure}
\vfill
\vspace{10pt}
\begin{subfigure}[h]{\linewidth}
\centering
\includegraphics[width=.85\linewidth]{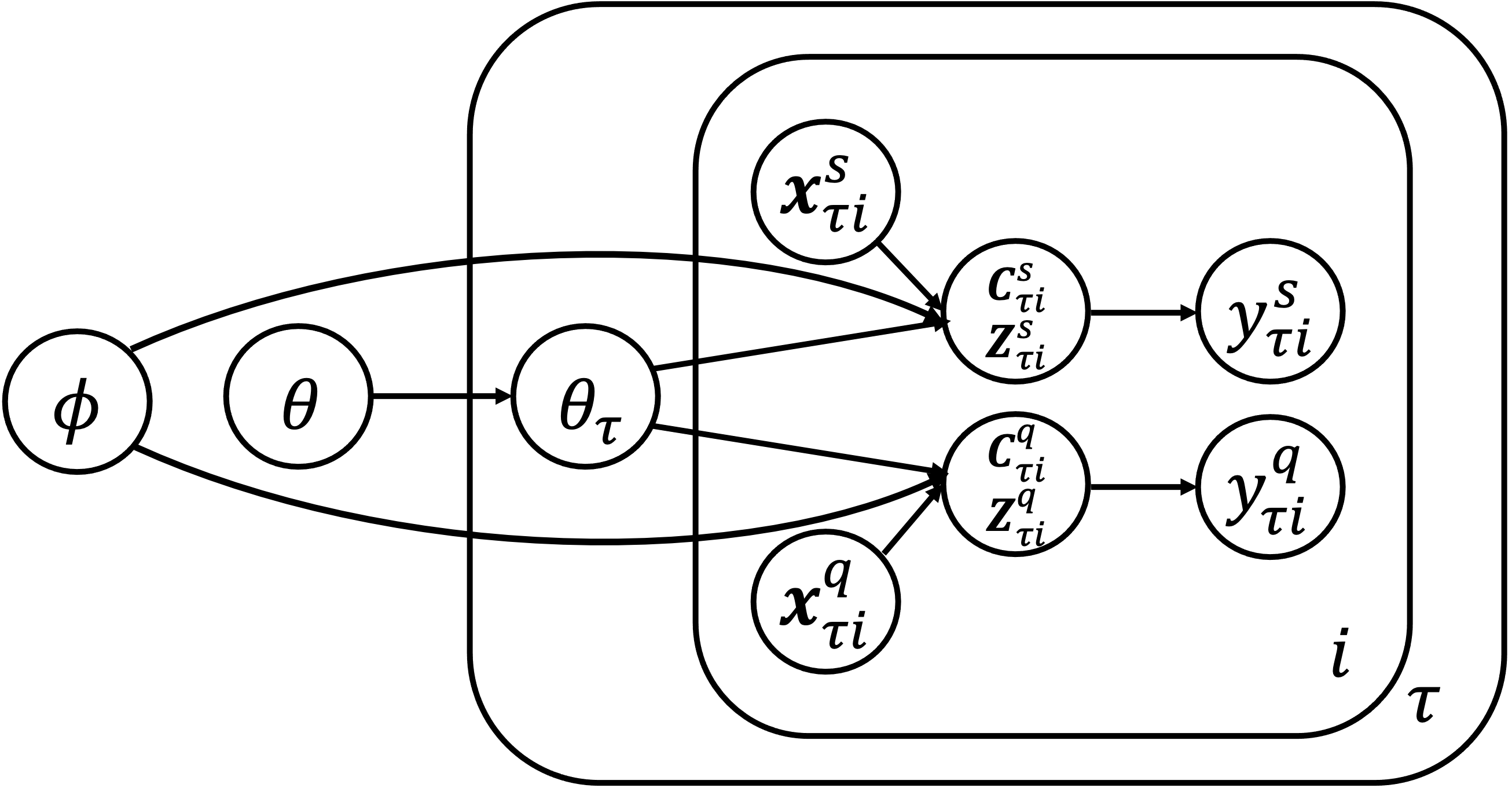}
\caption{The graphical model for concept inference}
\label{fig:graphical}
\end{subfigure}

\caption{An example illustrating the concepts in an image. The meaningful concepts are bounded by the red boxes in (a). The graphical model for our approach is depicted in (b), where $\phi$ is the parameters to generate the concept assignment $\mathbf{C}$ (red boxes in (a)).}
\label{fig:concept}
\end{figure}

However, it has been shown that current meta-learning methods are still way different from human beings' rapid learning process. Raghu et al. \cite{raghu2020rapid} demonstrate that current meta-learning methods are primarily doing feature reuse rather than rapid learning. Chen et al. \cite{chen2020closer} find that the advantage of the current meta-learning methods is  de facto tiny for a deep backbone. Furthermore, it is pointed out that human beings' ability to extract structural knowledge is still missing \cite{cao2021concept} in current methods. For example in Figure \ref{fig:bird}, human beings could easily extract the concepts of beak, wings, tail, and claw for the discrimination of different bird species, but such kind of concepts are not explicitly learned by current meta-learning algorithms. Therefore, we seek to design a meta-learning algorithm that explicitly extracts the structure from data.

There are lots of previous works in applying meta-learning to the structure discovery, most of which focus on the structure among the tasks \cite{liu2020adaptive, Jiang2019LearningTL, yao2019hierarchically, sohn2020meta, yao2020online}, i.e., whether two tasks share the same knowledge or the subtask structure. Some works \cite{Lake2011OneSL, doi:10.1126/science.aab3050, 7410499} also attempt to discover and utilize the feature structure by decomposing an image into multiple components. Recently, PDA-Net \cite{chen2021fewshot}  utilizes self-supervised learning to discover the local discriminative features for few-shot classification and Zhou et al. \cite{zhou2021metalearning} propose to meta-learn the symmetries by reparameterizing the convolutional neural networks. However, these studies on the feature structure only focus on the visual domain and are hard to generalize. The current understanding of the feature structure is shallow and the definition of the components in the structure  is still unclear from the machine learning perspective.

In this work, we focus on automatic feature structure discovery and its application to few-shot learning. Inspired by a recent work COMET \cite{cao2021concept}, which applies the human-defined concepts in few-shot learning, we  propose a new algorithm called \textit{Concept-Based Model-Agnostic Meta-Learning} (COMAML), which meta-learns the latent concepts behind data features without external labels. COMAML explicitly learns the concept assignment for each part of the data and regularizes it to a predefined concept distribution through variational inference. We illustrate COMAML's advantages in few-shot learning on a structured toy problem and further demonstrate its strengths on real-world datasets from biology and visual domains. The proposed COMAML has shown consistent improvements over previous few-shot learning methods and can sometimes even outperforms the method based on human-defined concepts. Qualitative examples also verify its great potential in concept discovery and model interpretability.

Our main contributions can be summarized as follows:
\begin{enumerate}
    \item We develop a concept discovery algorithm on top of the existing few-shot learning framework.
    \item We propose a new model that explicitly learns the concept assignment on data features.
    \item We demonstrate the advantage of our algorithm both quantitatively and qualitatively.
\end{enumerate}

\section{Problem Definition}

In this section, we will formally formulate the few-shot classification problem and briefly introduce the preliminaries of our method. We will stick to the following notation conventions in Table \ref{tab:symbols table} across the whole paper.

\begin{table}[ht]
     \centering
     \resizebox{\linewidth}{!}{
     \begin{tabular}{l|l}
         \toprule
         \textbf{Symbols}&\textbf{Definitions}\\
         \midrule
         $\tau$ & the task\\
         $\mathcal{S}_{\tau}$&the support set of task $\tau$ \\
         $\mathcal{Q}_{\tau}$&the query set of task $\tau$ \\
         $\mathbf{x}$ & the input data features \\
         $y$ & the class label \\
         $\mathbf{C}$ & the concept assignment associated with
         $\mathbf{x}$ \\
         $\mathbf{z}^{(j)}$&the $j$-th concept embedding\\
         $\xs, \xq$ & the set of data features in $\mathcal{S}_{\tau}$ and $\mathcal{Q}_{\tau}$ \\
         $\ys, \yq$ & the set of class labels in $\mathcal{S}_{\tau}$ and $\mathcal{Q}_{\tau}$ \\
         $\cs, \cq$ & the set of concept assignments in $\mathcal{S}_{\tau}$ and $\mathcal{Q}_{\tau}$ \\
         $\zs, \zq$ & the set of concept embeddings in $\mathcal{S}_{\tau}$ and $\mathcal{Q}_{\tau}$ \\
         $f_{\theta}$ & the feature encoder\\
         $h_{\phi}$ & the concept assignment generator\\
         $c$ & the number of latent concepts\\
         \midrule
        $\langle\cdot\rangle$ & the concatenation\\
        $\circ$ & the element-wise product \\
         \bottomrule
     \end{tabular}
     }
     \vspace{5pt}
     \caption{Symbols and notations.}
     \label{tab:symbols table}
 \end{table}

\subsection{Few-shot Classification}
Consider a distribution of tasks $p(\tau)$, each task $\tau=\{\mathcal{S}_{\tau}, \mathcal{Q}_{\tau}\}$, where $\mathcal{S}_{\tau}$ is the support set with a few labeled samples, and $\mathcal{Q}_{\tau}$ is an unlabeled query set sharing the same label space with the support set. For each data point $\{\mathbf{x}, y\}\in\mathcal{S}_{\tau}$, it consists of an input feature vector $\mathbf{x}\in\mathbb{R}^{d}$ and a class label $y\in\{1,\cdots,N\}$. Following the convention in previous works, we assume the number of samples from each class is balanced in the support set. Use $K$ to denote the number of samples per class in the support set, such a few-shot classification task is referred to as $N$-way $K$-shot classification. The label space of the training tasks and testing tasks is assumed to be disjoint. The goal of few-shot learning is to train a model that can generalize well on novel tasks. For each task, the model is evaluated by its classification accuracy on $\mathcal{Q}_{\tau}$ after the training on $\mathcal{S}_{\tau}$. 

\begin{figure*}
    \centering
    \includegraphics[width=0.7\textwidth]{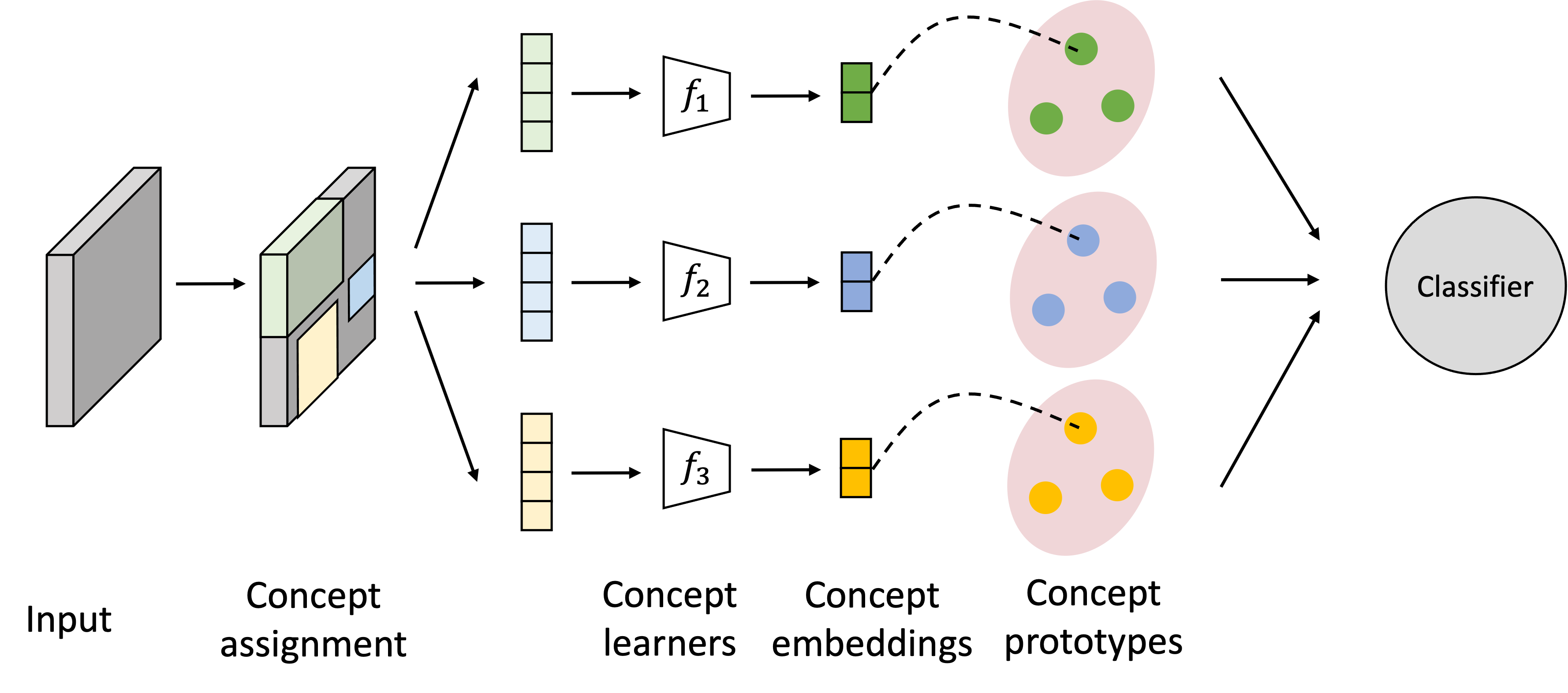}
    \caption{The overview of the proposed COMAML framework. COMAML consists of the concept assignment stage, concept encoders, and the classifier. Each dimension of the data features is first assigned to a concept, then the features belonging to the same concept will be encoded by the corresponding concept encoder into concept embeddings. They form the composite representation, which is passed to the classifier. The top classifier is initialized by the equivalent classification weights and bias of COMET from the concept prototypes. The concept encoders and the top classifier will get adapted on the support set through gradient descent. Best viewed in color.}
    \label{fig:architecture}
    \vspace{-10pt}
\end{figure*}
\subsection{Preliminaries} 
\paragraph{Prototypical Networks} 
Prototypical Networks \cite{snell2017prototypical} are a metric-based few-shot learning method that constructs a prototype for each class by projecting data features into an $m$-dimensional embedding space through the encoder $f_{\theta}: \mathbb{R}^{d}\rightarrow \mathbb{R}^{m}$. The prototype of class $k$  is defined as the the average of the embeddings of $k$'s support instances $\mathcal{S}^{k}_{\tau}$, 
\begin{equation}
    \mathbf{p}_{k} = \frac{1}{|\mathcal{S}^{k}_{\tau}|}\sum_{\{\mathbf{x}, y\}\in\mathcal{S}^{k}_{\tau}}f_{\theta}(\mathbf{x}).
\end{equation}
For classification, Prototypical Networks assign a new sample to the nearest prototype under the predefined distance measurement. The probability that a query sample $\mathbf{x}$ belonging to class $k$ could be calculated as

\begin{equation}
p(y=k|\mathbf{x}, \mathcal{S}_{\tau}) = \frac{\text{exp}(-d(f_{\theta}(\mathbf{x}), \mathbf{p}_k))}{\sum_{k'}\text{exp}(-d(f_{\theta}(\mathbf{x}), \mathbf{p}_{k'}))},
\end{equation}
where $d(\cdot,\cdot)$ is the distance function usually set as the squared Euclidean distance.

\paragraph{MAML}
Model-Agnostic Meta-Learning (MAML) \cite{finn2017modelagnostic} is a universal few-shot learning algorithm based on optimization. It tries to learn a good initialization for the model parameters that can be easily adapted to new tasks on limited labeled samples. The class probabilities output by MAML can be expressed as
\begin{align}
     p(y|\mathbf{x}, \mathcal{S}_{\tau}) &= \texttt{softmax}(\mathbf{b}_{\tau} + \mathbf{W}_{\tau} f_{\theta_{\tau}}(\mathbf{x})),\\
     (\mathbf{b}_{\tau}, \mathbf{W}_{\tau}, \pt) &= g(\mathbf{b}, \mathbf{W}, \theta, \mathcal{S}_{\tau}),
\end{align}
where $g$ corresponds to the adaptation rule such as the within-episode gradient descent. To assimilate the complementary strengths of Prototypical Networks and MAML, ProtoMAML \cite{triantafillou2020metadataset} initializes the linear classifier of MAML with the equivalent weights and bias used by Prototypical Networks. The use of a squared Euclidean distance in Prototypical Networks implies that the classification logits can be written as
\begin{equation}
\begin{split}
    -\|f_{\theta}(\mathbf{x})-\mathbf{p}_k\|^2 &= -f_{\theta}(\mathbf{x})^Tf_{\theta}(\mathbf{x}) + 2\mathbf{p}_k^Tf_{\theta}(\mathbf{x})-\mathbf{p}_k^T\mathbf{p}_k\\
                                        &= 2\mathbf{p}_k^Tf(\mathbf{x})-\|\mathbf{p}_k\|^2 + constant,
\end{split}
\end{equation}
where $constant$ is a class-independent scalar to be ignored. The $k$-th unit of the equivalent linear layer has weights $\mathbf{W}[k,:] = 2\mathbf{p}_k$ and bias $\mathbf{b}[k] = -\|\mathbf{p}_k\|^2$, which are both differentiable with respect to $\theta$.

\paragraph{Concept Learners for Meta-learning} COMET \cite{cao2021concept} is a few-shot learning method that utilizes the prior concept knowledge to boost the previous methods. It is grounded on Prototypical Networks and defines the prototypes for each concept. The concepts in COMET are defined as a set of binary masks $\{\mathbf{c}^{(j)}|\mathbf{c}^{(j)}\in\{0,1\}^d\}_{j=1}^c$, where each mask indicates the part of data belonging to that concept. The concept prototypes in COMET are defined as 
\begin{equation}
   \mathbf{p}_{k}^{(j)} = \frac{1}{|\mathcal{S}^{k}_{\tau}|}\sum_{\{\mathbf{x}, y\}\in\mathcal{S}^{k}_{\tau}}f_{\theta}^{(j)}(\mathbf{x}\circ \mathbf{c}^{(j)}),
\end{equation}
where $f_{\theta}^{(j)}(\cdot)$ is the concept encoder for the $j$-th concept. The class probability of class $k$ is accordingly calculated as 
\begin{align}
    p(y=k|\mathbf{x}, \mathcal{S}_{\tau}) = \frac{\text{exp}(-\sum_{j}d(f_{\theta}^{(j)}(\mathbf{x}\circ \mathbf{c}^{(j)}), \mathbf{p}_k^{(j)}))}{\sum_{k'}\text{exp}(-\sum_jd(f_{\theta}^{(j)}(\mathbf{x}\circ \mathbf{c}^{(j)}), \mathbf{p}_{k'}^{(j)}))}.
    \label{eq:comet}
\end{align}

\section{Methodology}
In this section, we will illustrate the inference algorithm on the latent concept assignment and further discuss the details of our proposed model.

\subsection{Inference on the Concept Assignment}
The objective of the standard few-shot learning on task $\tau$ can be formulated to maximize the log-likelihood of the objective $p(\yq|\xs,\ys,\xs)$. In this work, we treat the concept assignment $\mathbf{C}\in\{0,1\}^{d\times c}$ associated with the data features as latent variables, where $\mathbf{C}[:,j]$ is equivalent to $\mathbf{c}^{(j)}$ in COMET but we restrict $\mathbf{C}[i,:]$ to be one-hot. The graphical model is illustrated in Figure \ref{fig:graphical}. We focus on the inference for a specific task $\tau$ whose variational lower bound of the log-likelihood can be derived as
\begin{equation}
\begin{split}
    &\log{p(\yq|\xs,\ys,\xs)}\geq 
    \underset{\pt, \cq \sim q}{\mathbb{E}}[\log{p(\yq|\zq)} +\\ &\log{p(\cq|\xq)}+\log{p(\pt)}]+
    \mathcal{H}(q_{\phi}(\cq|\xq)) + \\
    &\mathcal{H}(q_{\phi}(\pt|\ys,\xs,\xq, \yq)).
\end{split}
\end{equation}
However, we face the challenge of estimating the posterior $q_{\phi}(\pt|\ys,\xs,\xq, \yq)$ since we do not have access to the labels $\yq$ during the test time. To simplify the variational inference procedure, we adopt the same strategy in Probabilistic MAML \cite{finn2019probabilistic} to use the maximum a posterior (MAP) value $\pt^*=\arg\max_{\theta_{\tau}} p(\theta_{\tau}|\xs,\ys)$. Therefore, the original variational inference objective could be rewritten as
\begin{equation}
\begin{aligned}
    &\log{p(\yq|\xs,\ys,\xs)}\geq 
    \underset{\cq \sim q}{\mathbb{E}}[\log{p(\yq|\zq)}+&\\ &\log{p(\cq|\xq)}]+
   \mathcal{H}(q_{\phi}(\cq|\xq))&\\
    & =\underset{\cq \sim q}{\mathbb{E}}[\log{p(\yq| \zq)}]&\textcolor{red}{(a)}\\&-D_{\text{KL}}(q_{\phi}(\cq|\xq)\| p(\cq|\xq)).&\textcolor{red}{(b)}
\end{aligned}
\end{equation}
Then we are left with two terms \textcolor{red}{(a)} and \textcolor{red}{(b)}, which correspond to the log-likelihood and regularization on $\cq$ respectively. 

For simplicity, we assume a uniform distribution for the prior $ p(\cq|\xq)$ in this work and we leave other choices for future exploration. The term \textcolor{red}{(b)} will degenerate to the entropy $\mathcal{H}(q_{\phi}(\cq|\xq))$, which prevents the trivial solution by assigning all features to the same concept.

The classification objective $L_{cls}$ is chosen to be the standard CrossEntropy loss and the final objective for the outer loop is the sum of both $L=L_{cls}-\mathcal{H}(q_{\phi}(\cq|\xq))$.

\subsection{Concept-based MAML}
Equipped with this inference objective,  we propose a Concept-based Model Agnostic Meta-Learning (COMAML) algorithm to meta-learn the concept assignment during the meta-training stage, the overview of which is shown in Figure \ref{fig:architecture}. Specifically, we denote the $j$-th concept embedding in COMAML as $\mathbf{z}^{(j)}=f_{\theta_{\tau}}^{(j)}(\mathbf{x}\circ \mathbf{C}[:,j])$.

We first combine COMET with MAML as the way ProtoMAML is developed. Using the squared Euclidean distance in Equation \ref{eq:comet}, we can express the $k$-th output logits of COMET as
\begin{equation}
\begin{split}
      &-\sum_j\|\mathbf{z}^{(j)}-\mathbf{p}_k^{(j)}\|^2\\=&-\sum_j {\mathbf{z}^{(j)}}^T\mathbf{z}^{(j)}+2\sum_j{\mathbf{p}^{(j)}_k}^T{\mathbf{z}^{(j)}}^T-\sum_j {\mathbf{p}^{(j)}_k}^T\mathbf{p}_k^{(j)}\\
      =&2(\langle\mathbf{p}^{(j)}_k\rangle_{j=1}^c)^T \mathbf{z}
      -(\langle\mathbf{p}^{(j)}_k\rangle_{j=1}^c)^T\langle\mathbf{p}^{(j)}_k\rangle_{j=1}^c+constant.
\end{split}
\label{eq:weight}
\end{equation}
 The $k$-th unit of the equivalent linear layer has weights $\mathbf{W}[k,:] = 2\langle\mathbf{p}^{(j)}_k\rangle_{j=1}^c$ and bias $\mathbf{b}[k]=\|\langle\mathbf{p}^{(j)}_k\rangle_{j=1}^c\|^2$.
Using this generalized form of COMET, we now discuss the details of the concept discovery process. 

Prior to the adaptation, we will generate all concept assignments $\mathcal{C}$ via a concept assignment generator $h_{\phi}(\mathbf{x})$.

For data with static semantics on each dimension, such as velocity and direction, the concept assignment generator could be directly parameterized by a probability matrix $\hat{\mathbf{C}}\in[0,1]^{d\times c}$. The concept assignment is sampled from a Gumbel-softmax distribution $\mathbf{C}[:,j]\sim\texttt{GS}(\hat{\mathbf{C}}[:,j], \lambda)$ with temperature $\lambda$. We anneal $\lambda$ towards $0$ during the training process, which makes the sampling more concentrated on the assignment with a high probability. In this case, the entropy is simple to calculate as we have direct access to probability distributions, which is the sum of entropies for concept assignment on each feature dimension $\mathcal{H}(\mathcal{C})=\frac{1}{|\mathcal{C}|}\sum_{\mathbf{C}\in\mathcal{C}}\sum_{i=1}^d\sum_{j=1}^c-\hat{\mathbf{C}}[i,j]\log{\hat{\mathbf{C}}[i,j]}$.

For data with dynamic semantics on each dimension like images, we make use of vector quantization technique \cite{Agustsson2017SofttoHardVQ, NIPS2017_7a98af17} to match patch features $pa(\mathbf{x})_{i}$ to a set of centroid vectors $\mathbf{e}_1, \cdots, \mathbf{e}_c$. The concept assignment is obtained as the index of the nearest centroid vector $\arg\min_j\|pa(\mathbf{x})_{i}-\mathbf{e}_j\|$. We add additional objectives as in VQ-VAE \cite{NIPS2017_7a98af17} to update the centroid vectors and patch features
\begin{equation}
    \|sg[pa(\mathbf{x})_i]-\mathbf{e}_j\|+\epsilon\|pa(\mathbf{x})_i-sg[\mathbf{e}_j]\|,
\end{equation}
where $sg$ stands for the stop-gradient operation which blocks the backpropagation, and $\epsilon$ is a scalar balancing the relative coefficient. These terms will be added to the previous objective $L$. In this case, we do not have direct access to probability distributions, so we use a surrogate distribution featured by probability $\frac{e^{-\|pa(\mathbf{x})_i-\mathbf{e}_j\|}}{\sum_{j'}e^{-\|pa(\mathbf{x})_i-\mathbf{e}_{j'}\|}}$ for $j$-th concept to compute the entropy.

With $\mathcal{C}$ in place, the initialized weights and bias can be generated according to Equation \ref{eq:weight}. During the inner loop, the parameters of the concept encoders $\theta_{\tau}$, weights $\mathbf{W}_{\tau}$ and bias $\mathbf{b}_{\tau}$ will get adapted. During the outer loop, COMAML will meta-learn both the initialization $\theta$ and concept assignment parameters $\phi$. The whole algorithm of COMAML is summarized in Algorithm \ref{alg:comaml}.
\begin{algorithm}[h]
\caption{COMAML}
\label{alg:comaml}
\begin{algorithmic}[1] 
\STATE \textbf{Input data}: $p(\tau)$\\
\STATE \textbf{Input parameters:} $\alpha$, $\beta$
\STATE Randomly initialize $\theta$, $\phi$
\FOR{iteration  $u=\{0,\cdots,U-1\}$}
\STATE Sample a batch of tasks $\mathcal{B}_u\sim p(\tau)$
\FOR{task $\tau\in\mathcal{B}_u$ }
\STATE $\theta^{(0)}\gets\theta$
\STATE Generate the assignment $\mathcal{C} \gets \{\mathbf{C}_{\mathbf{x}}| \mathbf{C}_{\mathbf{x}}=h_{\phi}(\mathbf{x}), (\mathbf{x},y)\in\mathcal{S}_{\tau}\}$ 
\STATE Generate the initialized weights and bias $\mathbf{W}^{(0)}$ and $\mathbf{b}^{(0)}$ according to Equation \ref{eq:weight}
\FOR{iteration $v\in\{0,\cdots,V-1\}$}
\STATE  Evaluate $l_{\tau}^s \gets L(f_{\theta^{(v)}}, \mathcal{C}, \mathbf{W}^{(v)}, \mathbf{b}^{(v)}, \mathcal{S}_{\tau})$ 
\STATE Update $\theta^{(v+1)} \gets \theta^{(v)} - \alpha \nabla_{\theta^{(v)}} l_{\tau}^s$ 
\STATE  Update $\mathbf{W}^{(v+1)} \gets \mathbf{W}^{(v)} - \alpha \nabla_{\mathbf{W}^{(v)}} l_{\tau}^s$
\STATE Update $\mathbf{b}^{(v+1)} \gets \mathbf{b}^{(v)} - \alpha \nabla_{\mathbf{b}^{(v)}} l_{\tau}^s$
\ENDFOR
\STATE Generate the assignment $\mathcal{C}\gets\{\mathbf{C}_{\mathbf{x}}|\mathbf{C}_{\mathbf{x}}=h_{\phi}(\mathbf{x}), (\mathbf{x},y)\in\mathcal{Q}_{\tau}\}$ 
\STATE Evaluate $l_{\tau}^q\gets L(f_{\theta^{(V)}},h_{\phi},\mathcal{C},\mathbf{W}^{(V)}, \mathbf{b}^{(V)},\mathcal{Q}_{\tau})$ 
\ENDFOR
\STATE Update $\theta\gets \theta - \beta \sum_{\tau\in \mathcal{B}_u}\nabla_{\theta}l_{\tau}^q$ 
\STATE Update $\phi\gets \phi- \beta \sum_{\tau\in \mathcal{B}_u}\nabla_{\phi}l_{\tau}^q$
\ENDFOR
\STATE\textbf{return} $\theta$, $\phi$
\end{algorithmic}
\end{algorithm}
\begin{table*}[h]
\centering
\begin{tabular}{ccccc}
\toprule
 & \multicolumn{2}{c}{\textbf{10 concepts}} &\multicolumn{2}{c}{\textbf{100 concepts}} \\
 \cmidrule(lr){2-3} \cmidrule(lr){4-5}
\textbf{Method}&\textbf{1-shot} & \textbf{5-shot} &\textbf{1-shot} & \textbf{5-shot}\\ 
\midrule
MAML & 60.8 $\pm$ 6.9 & 72.7 $\pm$ 3.7 & 74.8 $\pm$ 6.0 & 90.2 $\pm$ 2.1 \\
ProtoNet & 70.8 $\pm$ 5.1  & 87.2 $\pm$ 2.1 & 72.0 $\pm$ 6.4 &  86.8 $\pm$ 2.4 \\
ProtoMAML & 72.0 $\pm$ 5.9 & 89.6 $\pm$ 1.9 & 74.8 $\pm$ 5.5 & 91.8 $\pm$ 1.8 \\
\textbf{COMAML} & \textbf{78.8 $\pm$ 5.5} & \textbf{92.5 $\pm$ 1.6} & \textbf{82.0 $\pm$ 4.6} & \textbf{94.6 $\pm$ 1.6}\\
\midrule 
COMET$\dagger$ & 81.6 $\pm$ 4.0 & 92.7 $\pm$ 1.5 & 83.6 $\pm$ 4.1 & 95.2 $\pm$ 1.5\\
\bottomrule
\end{tabular}
\caption{Few-show classification results on the synthesized datasets. The best result on each dataset is bolded. $\dagger$ indicates the usage of the prior knowledge and not used in the comparison.}
\label{tab:toy}
  \vspace{-5pt}
\end{table*}

\section{Experiments}
In this section, we conduct the  empirical evaluation to answer the following research questions (RQ)
\begin{itemize}
    \item [  RQ1.] Can the concepts be discovered simply through optimization-based methods?
    \item [  RQ2.] How does COMAML compare with previous methods on real-world datasets?
    \item [  RQ3.] How sensitive is COMAML to the number of concepts?
\end{itemize}

\subsection{Classification on Structured Toy Problem}
Although COMET \cite{cao2021concept} has shown that the manual concept annotation could benefit few-shot learning as well, it remains unknown whether the concepts can be automatically discovered simply through optimization on deep neural networks. To answer RQ1, we design a toy problem where features can be split into different clusters. These kinds of data are in fact commonly seen in the real world such as the joints of the robot. 

The data generation process could be sketched below. We sample a set of latent vectors  $\{\mathbf{z}_i\}_{i=1}^n$ to form the class center $\langle\mathbf{z}_i\rangle_{i=1}^n$, where $n$ is the number of concepts and $\mathbf{z}_i\in\mathbb{R}^5$. Each dimension of $\mathbf{z}_i$ is uniformly sampled from the range $[-5,5]$. The class label is uniquely defined by the class center and the latent representation of a data point $\langle\mathbf{a}_i\rangle_{i=1}^n$ from this class is sampled from a normal distribution as $\mathbf{a}_i\sim\mathcal{N}(\mathbf{z}_i,0.2\cdot\mathbf{I})$, where $\mathbf{I}$ is the identity matrix. We define a set of fixed matrices $\{\mathbf{B}_i\}_{i=1}^{n}$ as the transformation from the latent representation to the observed data, where $\mathbf{B}_i\in\mathbb{R}^{5\times 30}$, and each element of $\mathbf{B}_i$ is also uniformly sampled from $[-5,5]$. Each latent vector $\mathbf{a}_i$ represents a concept and the corresponding matrices $\mathbf{B}_i$ only applies to this latent vector to generate the concept observation as $\sin(\mathbf{a}_i\mathbf{B}_i)$. Here $\sin(\cdot)$ serves as the non-linear transformation. The final observed data is the concatenation of all the observations of these concepts, i.e., $\langle\sin(\mathbf{a}_i\mathbf{B}_i)\rangle_{i=1}^n$. 

We compare COMAML with two optimization-based methods, MAML \cite{finn2017modelagnostic} and ProtoMAML \cite{triantafillou2020metadataset}, and one metric-based method, Prototypical Networks (ProtoNet) \cite{snell2017prototypical}. We also refer to COMET \cite{cao2021concept} with the ground-truth concept assignment as the theoretical upper bound of COMAML. We make $c$ in COMAML the same as the ground-truth number of concepts $n$ in the dataset. We use a two-layer MLP as the feature extractor with hidden dimensions $64$ and ReLU function for the non-linearity. For each layer, we also apply batch normalization and dropout with a rate of $0.2$. The optimization-based methods are optimized through a $1$-step gradient descent with a learning rate of $0.01$ in the inner loop and optimized by Adam  with a learning rate of $0.001$ in the outer loop. COMET and Prototypical Networks are optimized by Adam with a learning rate of $0.001$ as well. 

 The few-shot classification results are shown in Table \ref{tab:toy}, where we consider 5-way 1-shot and 5-way 5-shot classification tasks under $n=10$ and $n=100$.  It can be seen that COMAML has a clear advantage over the previous few-shot learning methods, with very close performance towards COMET using the ground-truth concepts. Although ProtoMAML achieves better performance than MAML and Prototypical Networks, COMAML still outperforms it by $7\%$ on 1-shot tasks and $3\%$ on 5-shot tasks in the final classification accuracy. The results demonstrate that concepts cannot be discovered simply through optimization, and it implies the necessity to introduce structure learning in the few-shot learning methods on the structured data.
\subsection{Classification on Real-world Datasets}
\begin{table*}[ht]
\centering
\begin{tabular}{ccccc}
\toprule
 & \multicolumn{2}{c}{\textbf{Tabula Muris}} &\multicolumn{2}{c}{\textbf{CUB}} \\
 \cmidrule(lr){2-3} \cmidrule(lr){4-5}

\textbf{Method}&\textbf{1-shot} & \textbf{5-shot} & \textbf{1-shot} & \textbf{5-shot} \\ 
\midrule
Finetune & 65.3 $\pm$ 1.0 & 82.1 $\pm$ 0.7 & 61.4 $\pm$ 1.0 & 80.2 $\pm$ 0.6 \\
MatchingNet & 71.0 $\pm$ 0.9 & 82.4 $\pm$ 0.7 & 61.0 $\pm$ 0.9  & 75.9 $\pm$ 0.6\\
MAML & 50.4 $\pm$ 1.1  & 57.4 $\pm$ 1.1 & 52.8 $\pm$ 1.0 & 74.4 $\pm$ 0.8    \\
RelationNet & 69.3 $\pm$ 1.0 & 80.1 $\pm$ 0.8 & 62.1 $\pm$ 1.0 & 78.6 $\pm$ 0.7  \\
MetaOptNet   & 73.6 $\pm$ 1.1 & 85.4 $\pm$ 0.9 & 62.2 $\pm$ 1.0 & 79.6 $\pm$ 0.6  \\
ProtoNet & 64.5 $\pm$ 1.0 & 82.5 $\pm$ 0.7 & 57.1 $\pm$ 1.0  & 76.1 $\pm$ 0.7 \\
DeepEMD  & NA & NA & 64.0 $\pm$ 1.0 & \textbf{81.1 $\pm$ 0.7} \\
ProtoMAML& 80.1 $\pm$ 1.0 & 89.5 $\pm$ 0.6 &63.0 $\pm$ 0.9&79.9 $\pm$ 0.61 \\
\textbf{COMAML} & \textbf{83.9 $\pm$ 0.9} & \textbf{93.6 $\pm$ 0.5} & \textbf{65.7 $\pm$ 1.0} & \textbf{81.1 $\pm$ 0.6} \\
\midrule
COMET$\dagger$ & 80.2 $\pm$ 0.9 & 92.6 $\pm$ 0.6 & 68.0 $\pm$ 1.0&85.6 $\pm$ 0.52\\
\bottomrule
\end{tabular}
\caption{Few-show classification results on  Tabula Muris and CUB datasets. The best result on each dataset is bolded. $\dagger$ indicates the usage of the prior knowledge and not used in the comparison. NA means the method is not applicable on the dataset.}
\label{tab:real}
  \vspace{-5pt}
\end{table*}

To answer RQ2, we compare COMAML with more few-shot learning methods on two real-world datasets from different domains.

\paragraph{Datasets.} We evaluate COMAML on two real-world datasets: Tabula Muris \cite{cao2021concept} and CUB-200 2011 \cite{WahCUB_200_2011}, which have static semantics and dynamic semantics on feature dimensions separately. The Tabula Muris dataset is a biology dataset for cell type classification, where the input features correspond to the gene expression profiles of cells. The CUB dataset is an image dataset for bird species classification. Similarly, we feed COMET with the human-defined concepts and make $c$ in COMAML the same as the number of human-defined concepts. 152 Gene Ontology \cite{article, osti_1581094} are used as the concepts for the Tabula Muris dataset and 21 part-based concepts such as beak, wing, tail, and claw are used for the CUB dataset.

\paragraph{Baselines.} For comparison, we select FineTune/Baseline++ \cite{chen2020closer}, Matching Networks (MatchingNet) \cite{vinyals2017matching}, MAML \cite{finn2017modelagnostic}, Relation Networks \cite{sung2018learning}, MetaOptNet \cite{lee2019metalearning}, DeepEMD \cite{zhang2022deepemd}, Prototypical Networks \cite{snell2017prototypical} and ProtoMAML \cite{triantafillou2020metadataset} as our baselines. We also use COMET \cite{cao2021concept} for reference, but since it utilizes additional prior knowledge, we do not directly compare COMAML with COMET. Besides, DeepEMD is only applicable to the image data so it has no results on the Tabula Muris dataset.

\paragraph{Implementation details.} The model architecture and the optimization are kept the same on the Tabula Muris dataset as on the synthesized datasets. On the CUB dataset, we apply a four-layer convolutional backbone as the feature extractor with input size $84\times 84$ across all methods. For COMAML, we add $c$ additional channels atop the last convolutional layer to generate the concept assignment for the $7\times 7$ feature maps/patch features. We simply use the elementwise max operation to combine all patch features belonging to the same concept into the concept embedding, without introducing any additional parameter except those centroid vectors for concept assignment. The optimization-based methods are optimized through a $5$-step gradient descent in the inner loop, and the learning rate is set as $0.01$. The Adam optimizer with a learning rate of $0.001$ is used in the outer loop and also the updates of other methods. 

\paragraph{Results.} The 5-way 1-shot and 5-way 5-shot classification results on these two datasets are shown in Table \ref{tab:real}. COMAML actually takes the lead on all tasks compared with baselines. Surprisingly, COMAML even outperforms COMET on the Tabula Muris dataset,  demonstrating the better quality of the learned concepts than the human-defined ones on it. In comparison with ProtoMAML, we can see that COMAML has a clear advantage with $4\%$ accuracy gain on Tabula Muris and $2\%$ accuracy gain on CUB, which is mainly brought by the concept discovery. The strongest baseline on CUB is DeepEMD, which achieves the same mean accuracy as COMAML on the 5-way 5-shot classification. DeepEMD makes use of the optimal transport to align different features for comparison, while COMAML uses a parameterized concept generator. Although COMAML does not show a clear advantage against DeepEMD on the 5-way 5-shot classification, it actually outperforms DeepEMD with $1.7\%$ in accuracy on the 5-way 1-shot classification. This is because the optimal transport is relatively inaccurate for the 1-shot case while the parameterized concept generator is meta-learned and insensitive to the number of samples in the task. Besides, DeepEMD can only align the features while COMAML also encourages clustering and extracts meaningful concepts for human interpretation. 

\subsection{Sensitivity Analysis}

\begin{figure}[h]
    \begin{minipage}{0.48\linewidth}
    \includegraphics[width=\linewidth]{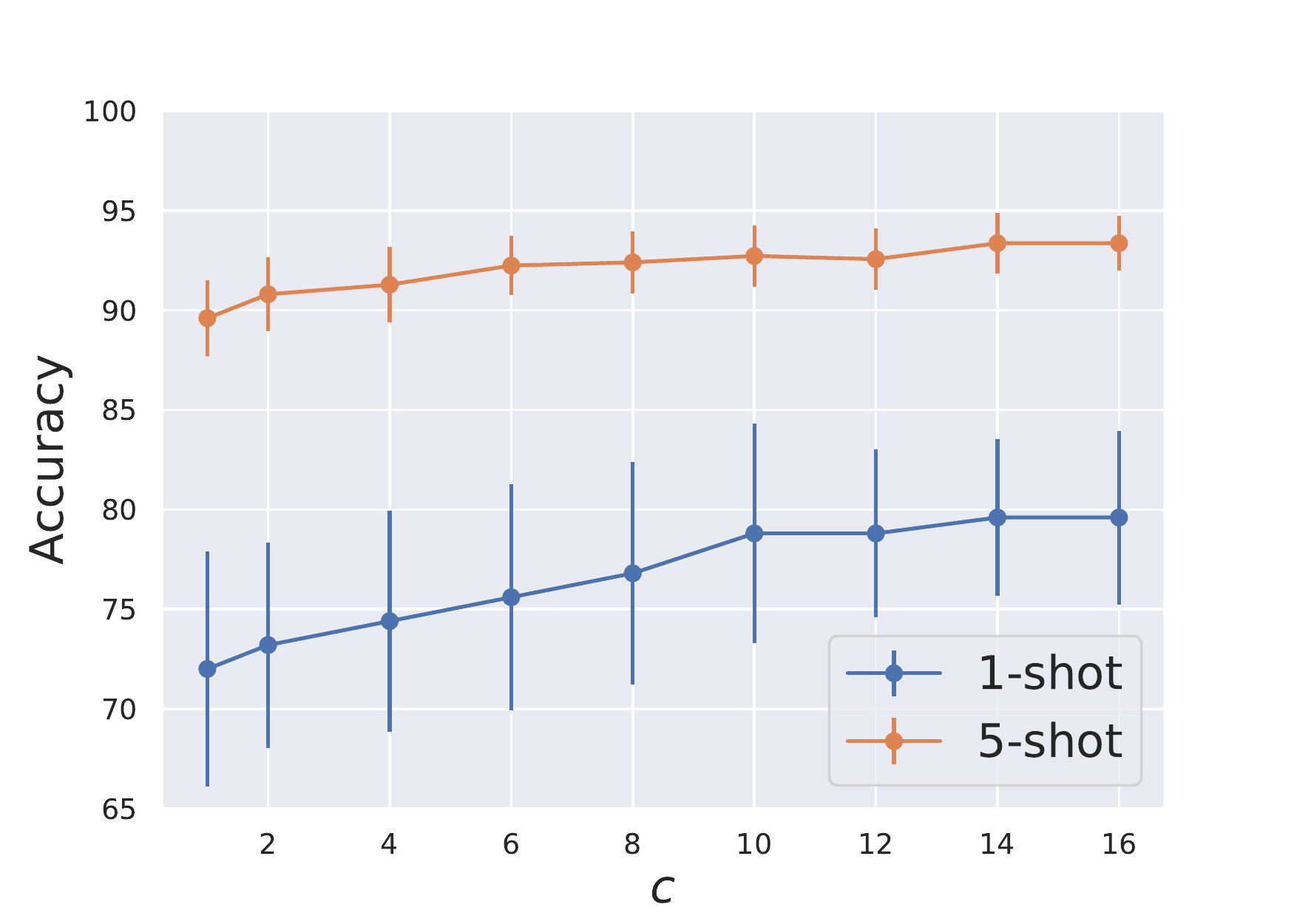}
    \end{minipage}
    \begin{minipage}{.48\linewidth}
    \includegraphics[width=\linewidth]{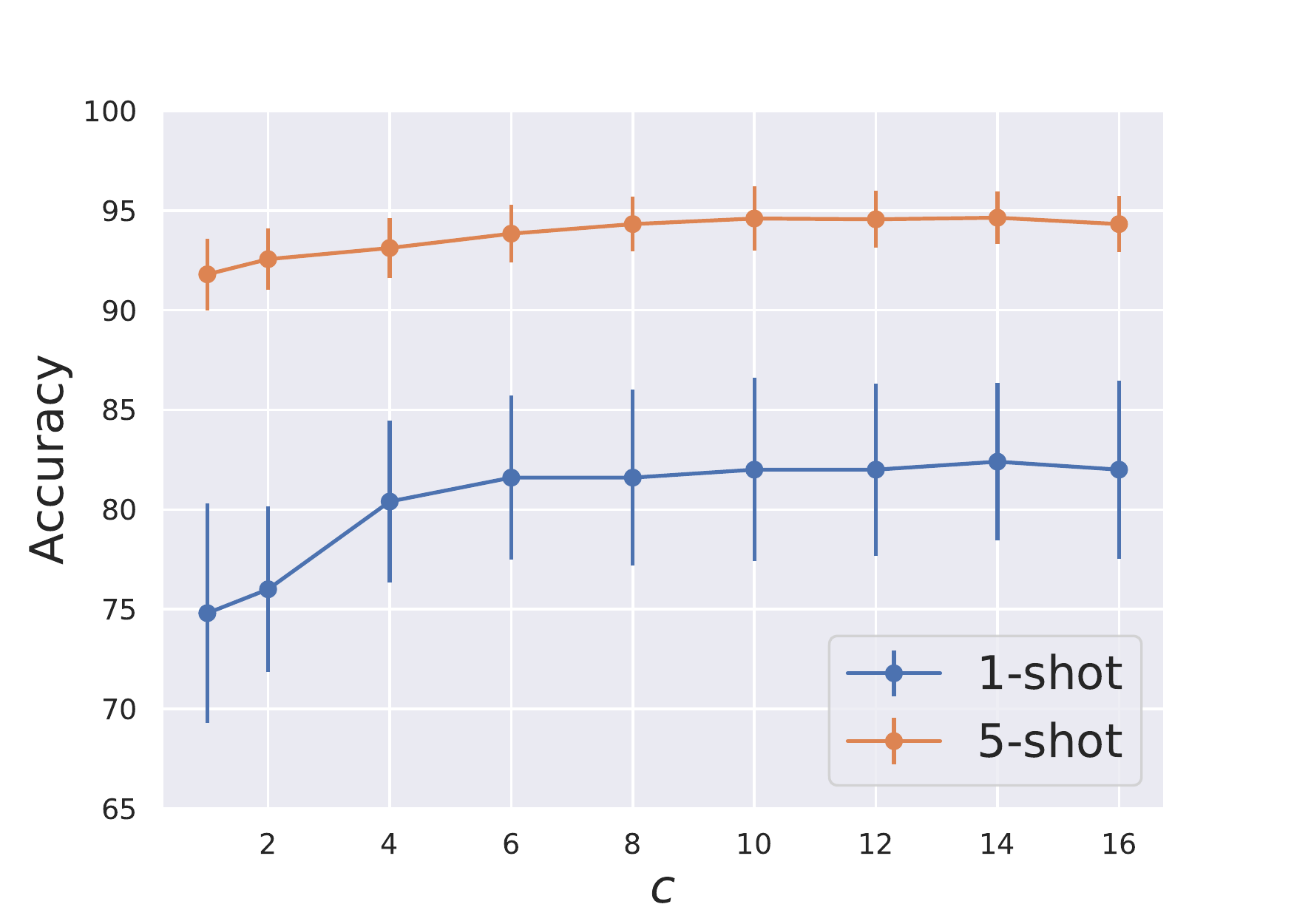}
    \end{minipage}
    \caption{Effect of the number of concepts on COMAML performance. The left figure shows the few-shot classification performance on the synthesized dataset with 10 concepts, and the right figure shows the few-shot classification performance on the synthesized dataset with 100 concepts.}
    \vspace{-5pt}
    \label{fig:syn_c}
\end{figure}

It is worth noting that COMAML is equivalent to ProtoMAML when the number of concepts is 1, so the choice of $c$ is also crucial to the final model performance. Since the actual number of concepts is unknown on the real-world datasets, we first investigate this problem on our synthesized datasets, and the results are visualized in Figure \ref{fig:syn_c}.

It can be seen that as the number of concepts increases, the performance becomes better and gradually reaches a steady state. No matter which case, introducing the concept discovery always boosts the performance. Furthermore, when the number of concepts of COMAML reaches the actual number of concepts, the model's performance is relatively stable and no negative effects appear when $c$ is too large, which demonstrates that COMAML is not sensitive to the choice of $c$.

\begin{figure}
    \centering
    \includegraphics[width=.7\linewidth]{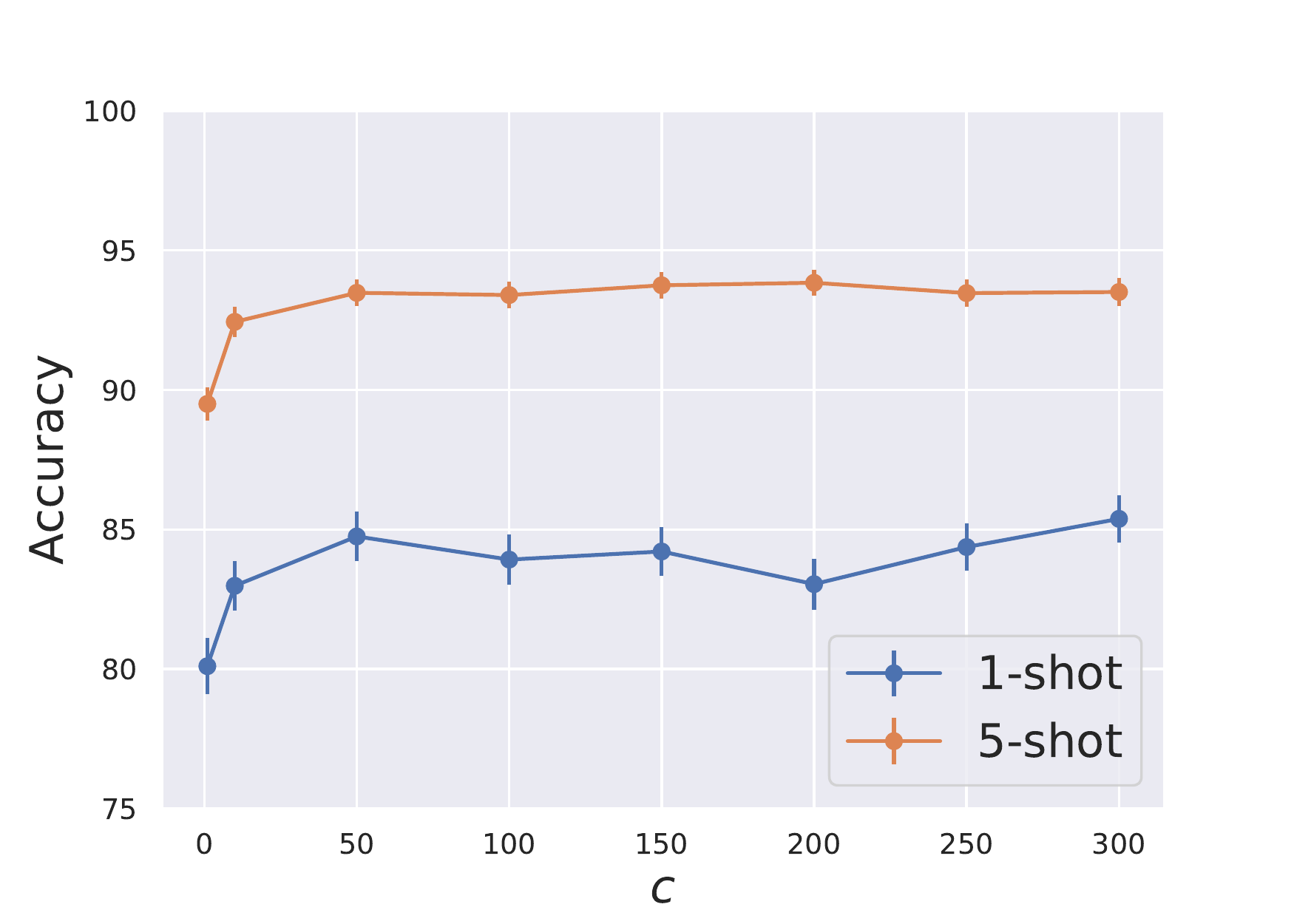}
    \caption{Effect of the number of concepts on COMAML performance on the Tabula Muris dataset.}
    \label{fig:TM_c}
    \vspace{-10pt}
\end{figure}

We then explore the effect of $c$ on the Tabula Muris dataset, where we present the results in Figure \ref{fig:TM_c}. 
Even by simply changing $c=1$ to $c=10$, COMAML has drastically increased the accuracy by $3\%$. And it shows a similar trend to stabilize when $c$ reaches 50. All the results above indicate that the tuning of COMAML on the number of concepts is very simple and introducing the concept discovery  is always beneficial to few-shot learning.

\subsection{Concept Visualization}
Finally, we present two examples to illustrate the concepts discovered by COMAML on the CUB dataset. Although there still exists some noise and redundancy, we find that COMAML indeed extracts some meaningful concepts from human beings' perspectives.

\begin{figure}[h]
\centering
    \begin{minipage}{0.48\linewidth}
    \centering
    \includegraphics[width=.95\linewidth]{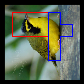}
    \end{minipage}
    \hfill
    \begin{minipage}{.48\linewidth}
     \centering
    \includegraphics[width=.95\linewidth]{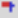}
    \end{minipage}
\vfill
\vspace{10pt}
    \begin{minipage}{0.48\linewidth}
     \centering
    \includegraphics[width=.95\linewidth]{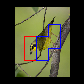}
    \end{minipage}
\hfill
    \begin{minipage}{.48\linewidth}
     \centering
    \includegraphics[width=.95\linewidth]{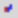}
    \end{minipage}
    \caption{Concepts extracted by COMAML. Red and blue colors correspond to two different concepts. The raw images are shown in the left column and the learned concepts are shown in the right column, whose corresponding positions are outlined in the raw images.}
    \label{fig:qual}
\end{figure}

In Figure \ref{fig:qual}, we visualize two concepts learned by COMAML, as annotated by the red color and blue color respectively. It can be seen that the concept annotated by the red color corresponds to the bird's head and the blue color represents the bird's body (including the belly and claw) in both images. These two qualitative examples demonstrate the power of COMAML in concept discovery and its potential for interpretability.

\section{Related Work}
In this section, we review the predominant few-shot learning methods and  related work in structured meta-learning.

\subsection{Few-shot Learning}
Few-shot learning is a prominent machine learning problem that aims to learn a new task through only a few samples. The current few-shot learning methods could be categorized into two main streams: the metric-based methods \cite{vinyals2017matching, snell2017prototypical, sung2018learning,gidaris2018dynamic, zhang2022deepemd} and the optimization-based methods \cite{finn2017modelagnostic, rusu2019metalearning,nichol2018firstorder, grant2018recasting, antoniou2019train, triantafillou2020metadataset}.
The metric-based methods aim to find a mapping to the space where the instances of the same class are close to each other and are characterized by different similarity measurements.  For example, Matching Networks \cite{vinyals2017matching} use the cosine similarity,  Prototypical Networks \cite{snell2017prototypical} choose the negative distance to the prototypes, while  Relation Networks \cite{sung2018learning} learn a parameterized `relational module'. Some works \cite{hou2019cross,10.5555/3045390.3045585} also propose to apply the attention mechanism to the local discriminative features. Recently, DeepEMD \cite{zhang2022deepemd} innovatively utilizes the Earth Mover's distance \cite{Rubner2004TheEM} to measure the optimal distance among image features, where patch features from different images are assigned to compare based on the optimal transport. By contrast, COMAML generates the assignment based on a parameterized function which can be learned as meta-knowledge. The optimization-based methods aim to achieve a good generalization on new tasks with a few fine-tuning steps on limited labeled samples. Model-Agnostic Meta-Learning (MAML) \cite{finn2017modelagnostic} is first proposed to learn a model initialization that can be easily adapted for any new task through gradient descent. On top of this universal framework, many follow-up works extend MAML to tackle its limitations under different scenarios. To deal with the task ambiguity, Kim et al. \cite{kim2018bayesian} and Finn et al. \cite{finn2019probabilistic} extend MAML to the probabilistic version through variational inference. Meta-Transfer Learning (MTL) \cite{sun2019metatransfer} combines transfer learning and meta-learning to reduce the risk of overfitting. In light of the difficulties when operating high-dimensional parameters space on a low-data regime, Latent Embedding Optimization (LEO) \cite{rusu2019metalearning} introduces a generation process of the model parameters and performs gradient-based adaption on the low-dimensional space. To assimilate the complementary benefits of the metric-based methods and optimization-based methods, ProtoMAML \cite{triantafillou2020metadataset} uses an equivalent linear layer to the Prototypical Network as the initialization of the top linear classifier in MAML. In this work, COMAML can be treated as a generalized form of ProtoMAML with composite features taken into account. 

 \subsection{Structured Meta-learning }
 Here we use the general term \textit{structured meta-learning} to refer to the meta-learning algorithms which learn or use the structure (prior knowledge) as the meta-knowledge. We categorize these methods according to the structure they utilize, basically the task structure and the feature structure. The task structure specifies the relationship among tasks, which can be used to decompose a new task as a combination of previously seen tasks. Liu et al. \cite{liu2020adaptive} and Jiang et al. \cite{Jiang2019LearningTL} focus on the relationship among classes, where the classification of a new class get benefits from learning on those strongly dependent classes. Hierarchical Structured Meta-Learning (HSML) ~\cite{yao2019hierarchically} and Online Structured Meta-learning (OSML) \cite{yao2020online} decompose each task into sub-modules and extracts the pathway over them for a new task before the adaptation. 
 
 The feature structure indicates the distribution of the data, which in turn informs a more proper model architecture. In the earliest trial, Luke et al. \cite{Lake2011OneSL, doi:10.1126/science.aab3050} apply  Bayesian probabilistic programs on the pre-defined individual strokes for the hand-written character recognition task. The follow-up work \cite{7410499} extends this method by replacing the manually defined features with object descriptors on the image classification.  Tokmakov et al. \cite{https://doi.org/10.48550/arxiv.1812.09213} uses the human-defined labels to regularize the compositionality of the learned representation and COMET \cite{cao2021concept} uses separate encoders for each concept.  Besides these composite structures, Chen et al. \cite{chen2021fewshot} use a self-supervised method to discover the discriminative parts for the few-shot learning purpose, and Zhou et al. \cite{zhou2021metalearning} propose to meta-learn the equivariances from data by reparameterizing the convolutional neural networks. Our method COMAML also focuses on the feature structure, which is based on COMET. However, COMAML automatically learns the composite structure via meta-learning and is a more universal framework for feature structure learning not limited to visual data.

\section{Conclusion}
In this paper, we propose a simple yet effective method to realize few-shot learning on structured data by introducing a concept discovery module atop the existing few-shot learning model. The proposed algorithm, COMAML, decomposes the data features into different concepts, applies separate encoders and regularizes the concept assignments via variational inference. It achieves consistently better performance on both the synthesized and real-world datasets, sometimes even outperforming the model utilizing the human-defined concepts. COMAML is inspired by structured knowledge from human beings and can be widely applied to various real-world tasks. In the future, we plan to further investigate the following aspects, (1) extend COMAML to more domains like text and graphs, (2) take into account the uncertainty from partially-observed concepts, and (3) experiment with a broad family of prior distributions for the concept assignment and composite representation.

\section*{Acknowledgement}

This work is partially supported by NSF (1947135, 
2134079 
and 1939725 
), DARPA (HR001121C0165), NIFA (2020-67021-32799) and ARO (W911NF2110088).

\bibliographystyle{acm}
\bibliography{reference}
\end{document}